\documentclass[letterpaper,twocolumn]{article}
\usepackage{isairas}

\usepackage{cite}
\usepackage{amssymb,amsfonts}
\usepackage{algorithmic}
\usepackage{graphicx}
\usepackage{textcomp}

\usepackage[usenames, dvipsnames,table]{xcolor}
\usepackage{subcaption}
\usepackage[colorinlistoftodos]{todonotes}

\usepackage{graphicx, import}

\usepackage{algorithmic}
\usepackage{algorithm}

\usepackage{siunitx}

\usepackage[capitalise]{cleveref}

\renewcommand{\baselinestretch}{0.92}

\begin{document}

\definecolor{lightgray}{gray}{0.9}

\newcommand{\eg}{{e.g.}}
\newcommand{\ie}{{i.e.}}

\newcommand\mydots{\makebox[1em][c]{.\hfil.\hfil.}}
\newcommand{\aff}{assistive free-flyer}
\newcommand{\dof}{DoF}
\newcommand{\tof}{ToF}
\newcommand{\mlp}{MLP}
\newcommand{\pac}{PAC}

\newcommand{\ac}[1]{\textcolor{MidnightBlue}{\textrm{(#1 - AC)}}}
\newcommand{\acmargin}[2]{{\color{MidnightBlue}#1}\marginpar{\color{MidnightBlue}\raggedright\footnotesize [AC]: #2}}

\newcommand{\tc}[1]{\textcolor{purple}{\textrm{(#1  - TC)}}}
\newcommand{\tcmargin}[2]{{\color{purple}#1}\marginpar{\color{purple}\raggedright\footnotesize [TC]: #2}}

\newcommand{\sas}[1]{\textcolor{Aquamarine}{\textrm{(#1  - SAS)}}}
\newcommand{\sasmargin}[2]{{\color{Aquamarine}#1}\marginpar{\color{Aquamarine}\raggedright\footnotesize [SAS]: #2}}

\newcommand{\md}[1]{\textcolor{PineGreen}{\textrm{(#1 - MD)}}}
\newcommand{\mdmargin}[2]{{\color{PineGreen}#1}\marginpar{\color{PineGreen}\raggedright\footnotesize [MD]: #2}}

\newcommand{\rbr}[1]{\textcolor{RubineRed}{\textrm{(#1 - RBR)}}}
\newcommand{\rbrmargin}[2]{{\color{RubineRed}#1}\marginpar{\color{RubineRed}\raggedright\footnotesize [RBR]: #2}}

\newcommand{\amv}[1]{\textcolor{RoyalPurple}{\textrm{(#1 - AMV)}}}
\newcommand{\amvmargin}[2]{{\color{RoyalPurple}#1}\marginpar{\color{RoyalPurple}\raggedright\footnotesize [AMV]: #2}}

\newcommand{\mpv}[1]{\textcolor{BurntOrange}{\textrm{(#1 - MPV)}}}
\newcommand{\mpvmargin}[2]{{\color{BurntOrange}#1}\marginpar{\color{BurntOrange}\raggedright\footnotesize [MPV]: #2}}

\newcommand{\mc}[1]{\textcolor{Plum}{\textrm{(#1 - MC)}}}
\newcommand{\mcmargin}[2]{{\color{Plum}#1}\marginpar{\color{Plum}\raggedright\footnotesize [MC]: #2}}

\newcommand{\ab}[1]{\textcolor{RubineRed}{\textrm{(#1 - AB)}}}
\newcommand{\abmargin}[2]{{\color{RubineRed}#1}\marginpar{\color{RubineRed}\raggedright\footnotesize [AB]: #2}}   

\title{\LARGE \bf
Design and Development of a Gecko-Adhesive Gripper for the Astrobee Free-Flying Robot
}

\author[1]{Abhishek Cauligi$^*$}
\author[1]{Tony G. Chen$^*$}
\author[1]{Srinivasan A. Suresh}
\author[2]{Michael Dille}
\author[2]{Ruben Garcia Ruiz}
\author[2]{Andres Mora Vargas}
\author[1]{Marco Pavone}
\author[1]{Mark R. Cutkosky}
\affil[1]{Stanford University, 450 Jane Stanford Way, Stanford, CA 94305, USA}
\affil[2]{Intelligent Robotics and Astrobee Facilities Groups, NASA Ames Research Center, Moffett Field, CA 94043}


\affil[ ]{E-mails: \{ \tt acauligi, agchen, sasuresh, pavone, cutkosky\}{\tt @stanford.edu}, \{\tt michael.dille, andres.moravargas, ruben.m.garciaruiz\} {\tt@nasa.gov}}
\affil[ ]{$^*$These authors contributed equally to this work.}

\maketitle
\thispagestyle{empty}

\begin{abstract}
Assistive free-flying robots are a promising platform for supporting and working alongside astronauts in carrying out tasks that require interaction with the environment. 
However, current free-flying robot platforms are limited by existing manipulation technologies in being able to grasp and manipulate surrounding objects. 
Instead, gecko-inspired adhesives offer many advantages for an alternate grasping and manipulation paradigm for use in assistive free-flyer applications. 
In this work, we present the design of a gecko-inspired adhesive gripper for performing perching and grasping maneuvers for the Astrobee robot, a free-flying robot currently operating on-board the International Space Station. 
We present software and hardware integration details for the gripper units that were launched to the International Space Station in 2019 for in-flight experiments with Astrobee. 
Finally, we present preliminary results for on-ground experiments conducted with the gripper and Astrobee on a free-floating spacecraft test bed.
\end{abstract}

\section{Introduction}
Assistive free-flying robots have received significant interest in recent years due to their promise in augmenting and enhancing the capabilities of astronauts performing tasks in space. 
In addition to hands-free video and audio capabilities, \aff{}s provide the ability to grasp and manipulate objects during intravehicular (IVA) and extravehicular (EVA) activities. 
For example, an \aff{} could be tasked with retrieving a tool from a particular module on the International Space Station (ISS) or carrying out repairs outside the spacecraft, thereby freeing hours of astronaut time spent conducting menial tasks and reducing or eliminating higher risk tasks for astronauts. 
Key to the success and versatility of such \aff{}s is the ability of the robots to perform tasks requiring physical interaction with both the grasp target and environment. 
Assistive free-flyers must be equipped with interaction tools that can grasp and manipulate objects around them, a task that can be particularly challenging to execute in microgravity. 

However, traditional robotic grippers, which rely on pinching (\ie{}, force closure) or caging (\ie{}, form closure), often require targeting a small set of specially pre-defined designated features and need precise positioning with minimal relative velocities to succeed. 
To overcome these shortcomings, gecko-inspired adhesives are being explored as an alternate grasping and manipulation paradigm with significant advantages. 
They provide several capabilities that make them promising for both IVA and EVA activities:
\begin{enumerate}
    \item \emph{Versatility}: gecko adhesives do not require enclosing the features of an object (form closure) or pinching on opposing faces of an object (force closure).
    \item \emph{Controllability}: gecko adhesives can remain attached to a surface without any power draw and impart little to zero momentum when grasping and releasing an object. 
    \item \emph{Dynamic}: gecko adhesives can grasp translating and spinning objects, thereby extending their use to capturing non-cooperative or tumbling objects~\cite{EstradaHockmanEtAl2016}.
\end{enumerate}


\begin{figure}[t!]
\centering
\includegraphics[width=.35\textwidth]{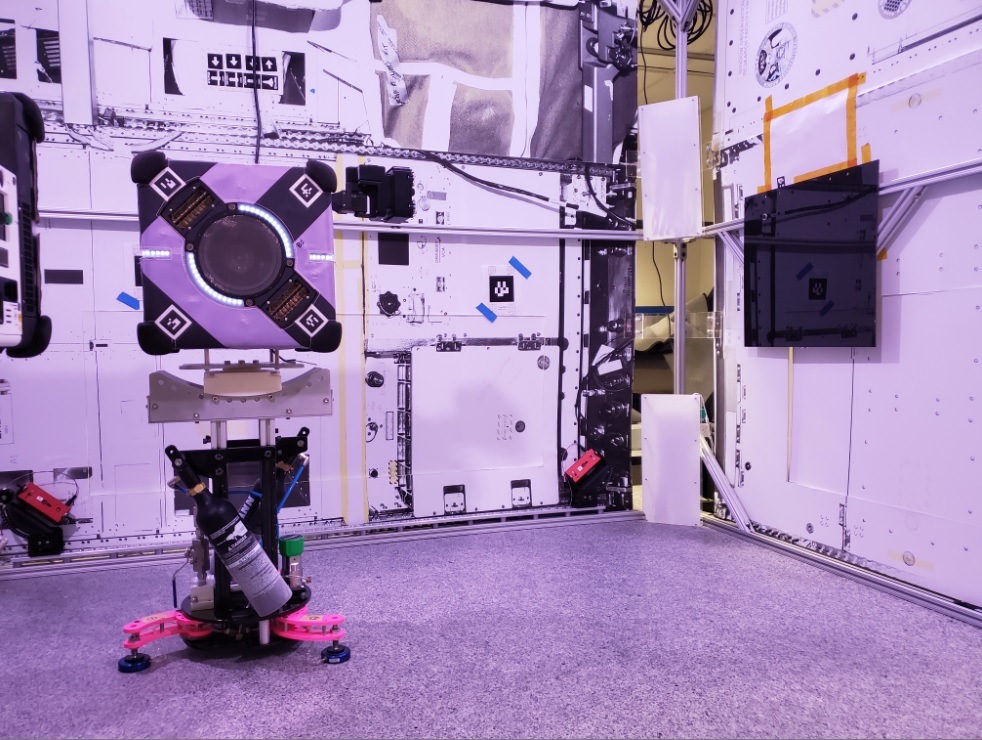}
\caption{Perching experiments conducted with the flight gecko gripper integrated onto an Astrobeee unit at NASA Ames Research Center granite table facility~\cite{Vargas2018}.}
\label{fig:granite_table}
\end{figure}

While gecko-inspired adhesion has several promising applications such as satellite servicing~\cite{MacPhersonHockmanEtAl2017,Flores-AbadMaEtAl2014} and space debris capture~\cite{BylardMacPhersonEtAl2017}, in this work we focus our attention on the application of gecko-inspired grasping for \aff{}s. 
Specifically, we present the design and development of a flight-qualified gecko-adhesive gripper for the Astrobee free-flying robot, a guest science platform used for technology demonstration purposes on the ISS~\cite{BualatSmithEtAl2018}. 
Details on the flight-certification of a gecko-adhesive gripper for on orbit experiments with the Astrobee and the associated mechanical, avionics, and software integration are discussed. 
Further, we provide details on planned activities in 2021 with the two flight grippers that were launched to the ISS in 2019.

\section{Background}
\subsection{Astrobee Free-Flying Robot}
The excessive cost and associated safety considerations of crewed missions are often a prohibitive factor in carrying out tasks in space such as satellite servicing (\eg{} repair and capture) or managing orbital debris. 
Assistive free-flying robots are a promising platform that can greatly help to reduce time and effort spent by astronauts in carrying out repetitive or unsafe tasks. 
For example, these robots could both work with or without astronaut supervision, perform various logistics, monitoring, and maintenance tasks, and can thereby save valuable crew time on-board the ISS.

One of the first such free-flying platform programs was the SPHERES Guest Science Program~\cite{Enright2004} that demonstrated the capabilities of assistive free-flyers for autonomous operations on the ISS. 
However, the SPHERES robots were encumbered by limited computational resources, cold gas thrusters requiring continuous maintenance by astronauts, and a specialized software development procedure. 
To address these shortcomings, the Astrobee robot~\cite{SmithBarlowEtAl2016,BualatSmithEtAl2018} was developed as a successor to the SPHERES platform. 
Instead of cold gas thrusters, Astrobee provides holonomic motion along six degrees-of-freedom (\dof{}) using battery-powered fans and autonomously docks to a custom charging station for recharging batteries. 
Further, the Astrobee flight software builds upon the Robot Operating System (ROS), an open-source framework that allows for modular and rapid software development. 

Astrobee is further equipped with a 3-\dof{} perching arm with joint control managed by the perching arm controller (\pac{}) board~\cite{Park2017}. 
The perching arm uses an underactuated claw gripper to perch on the ISS handrails, but this gripper is limited to use on target objects with pre-defined features due to the reliance of the design on ``pinching'' or ``caging'' grasps. 
The target object is also assumed to be stationary relative to the gripper, an assumption that further inhibits the applicability of the gripper to the versatile set of tasks that may arise on a daily basis. 
Instead, the paradigm of gecko-inspired adhesives show promise in addressing these limitations and are the focus of attention in this work. 

\subsection{Gecko-inspired Adhesives}


Gecko-inspired adhesives draw inspiration from the micro- and nano-structures found on the toes of geckos. 
Each gecko toe is comprised of arrays of hundreds of thousands of microscopic structures called \emph{setae} that split into hundreds of \SI{200}{\nano\meter} wide spatular tips ~\cite{Autumn2006}. 
This hierarchical structure collapses when force is applied to it in the shear direction, becoming uniformly flush with the substrate and maximizing their contact area. 
van der Waals forces then induce adhesion between these structures and the substrate. 
Further, this adhesion is highly directional, \ie{} when loaded in the adhesive preferred shear direction, they exhibit strong adhesion while allowing low force release when other directional shear forces are introduced.

\begin{figure}[t!]
	\centering
	\includegraphics[width=0.95\columnwidth]{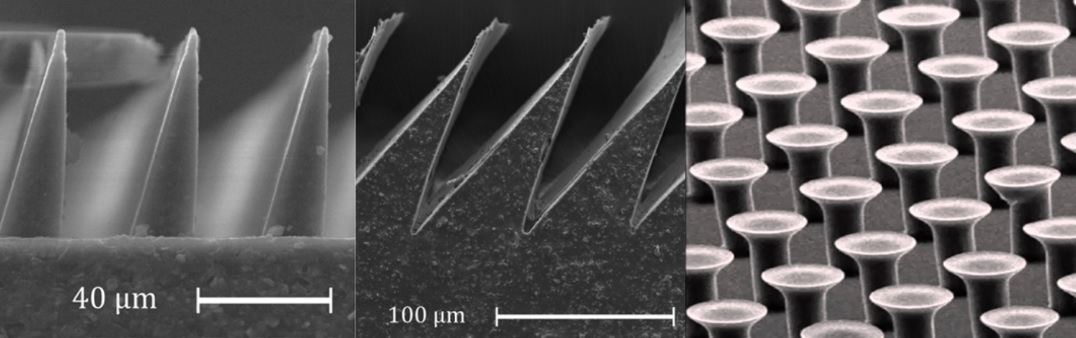}
	\caption{Micro-wedge-shaped gecko-inspired adhesives shown on the left and center. Mushroom-shaped gecko-inspired adhesives shown on the right~\cite{Autumn2006,Song2017}. }
	\label{fig:gecko_adhesives}
\end{figure}

Motivated by these desirable properties of the natural gecko setae structures for adhesion, synthetic gecko-inspired adhesives have been manufactured using the same principles to generate adhesion on relatively smooth surfaces. 
As shown in~\cref{fig:gecko_adhesives}, there are two broad classes of synthetic gecko adhesives. 
The first class consists of adhesives using stalks of soft polymer with mushroom-shaped tips~\cite{Song2017,Trentlage2018,Trentlage2018a}. 
Such mushroom-shaped tips provide strong adhesion, but unlike the natural gecko setae they draw inspiration from, they require high forces to engage and disengage the adhesives. 
For grasping and manipulation tasks in a microgravity environment, this could lead to the adhesives imparting non-negligible force upon adhering and releasing, disturbing the grasped object. 
Thus, such mushroom-shaped tips are less suitable for use in manipulating objects in a micro- or zero-gravity environment.


The second class of synthetic adhesives consists of wedge-shaped gecko-inspired adhesives. 
This class of adhesives allows for rapid attachment and detachment with minimal interaction forces between the adhesion and grasped surface. 
Further, the wedge-shaped adhesives can be reused and maintain adhesion up to at least 30,000 loading cycles, have been manufactured using space-grade, low-volatility silicones, and have been tested in wide temperature ranges, vacuum environment, and under radiation. The light-touch properties and the proven environmental compatibility drove the selection of the wedge-shaped adhesives for the gripper presented in this work.


Recently, gecko-inspired adhesives have emerged as an exciting tool for building grippers for various spacecraft robotic applications. 
In~\cite{JiangHawkesEtAl2017}, the efficacy of gecko-inspired adhesives was first demonstrated for robotics applications, and a flat surface gripper was used to enable a large, four-legged robot to climb solar panels. 
A curved surface gripper was built to enable a free-flying robot to grasp uncooperative, spinning targets~\cite{MacPhersonHockmanEtAl2017,EstradaHockmanEtAl2016}, but these experiments were limited to a 3-\dof{} granite table test bed. 
Jiang, et al. presented results from micro-gravity experiments conducted with gecko-inspired adhesives, where a hand-held gripper was constructed to capture and release large flat and curved objects during a parabolic flight test~\cite{JiangHawkesEtAl2017}. 
Manual testing with a spring loaded gripper unit was conducted on the ISS, but these experiments did not use an assistive free-flying robot platform~\cite{Parness2017a}. 
Thus, we seek to bridge this gap by equipping a free-flying robot with gecko-inspired adhesives on the ISS to investigate the efficacy of this technology for dexterous grasping and manipulation.

\begin{figure}[h]
\centering
\includegraphics[width=0.95\columnwidth]{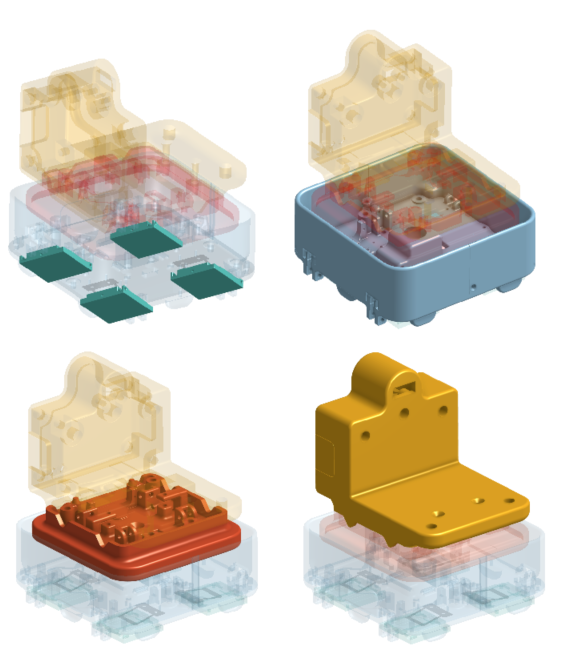}
\caption{Four different sub-assemblies of the gecko-inspired adhesive gripper: adhesive tiles in teal, palm assembly in light blue, wrist assembly in orange and base assembly in yellow.} 
\label{fig:gripper_section}
\end{figure}

\begin{figure}[t]
\centering
\includegraphics[width=.5\textwidth]{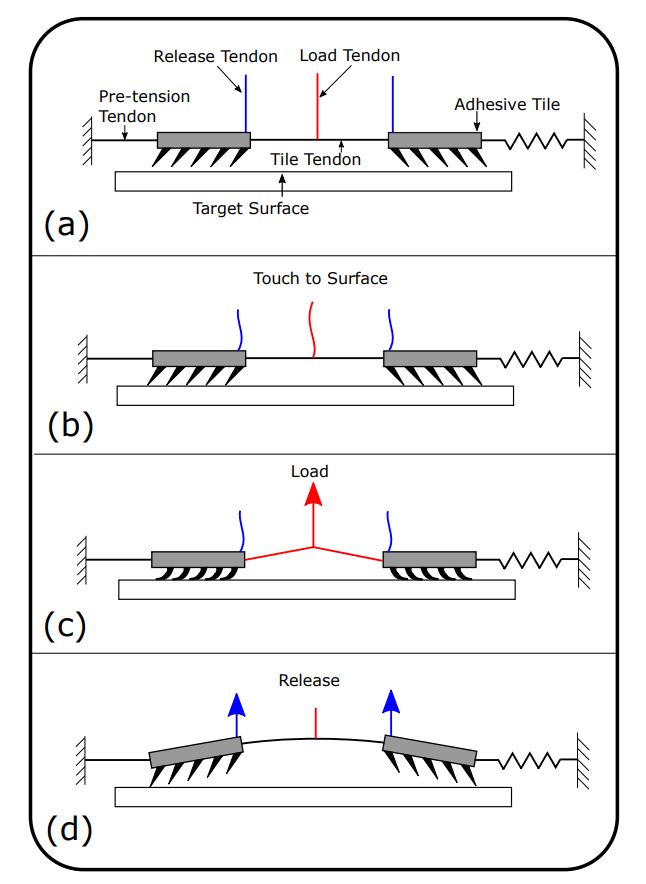}
\caption{Schematic edge-on view of an adhesive tile pair indicating tendon topology and function. A combination of tendons are used to load the tiles to generate useful grasp forces. Upon release, the tendons allow for a nearly zero-force release of the grasped object from the surface. Figure adapted from~\cite{Hawkes2015}.}.
\label{fig:loading_sequence}
\end{figure}


\section{Gripper Design}
As the flight unit of the gecko-inspired adhesive gripper is intended to serve as a drop-in replacement for the end-effector of the existing Astrobee perching arm, particular care was taken to ensure mechanical and electrical compatibility of the developed flight gripper and the Astrobee robot. 
This section will detail the mechanical, avionics, and software design considerations involved for the flight gripper.

\begin{figure*}[t!]
\centering
    \begin{subfigure}[t]{0.2\textwidth}
  \centering
  \captionsetup{justification=centering}
    \includegraphics[width=1.75in,]{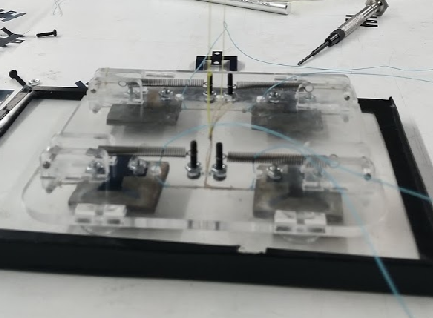}
    \caption{}
    \label{fig:design_iteration1}
  \end{subfigure}
  \qquad
  \begin{subfigure}[t]{0.2\textwidth}
  \centering
  \captionsetup{justification=centering}
    \includegraphics[width=1.75in,]{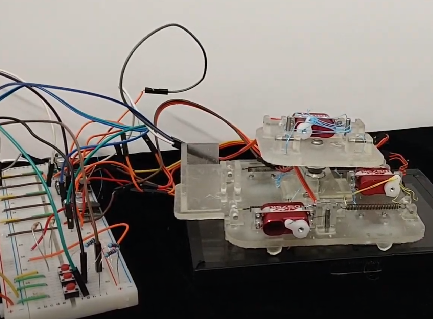}
    \caption{}
    \label{fig:design_iteration2}
  \end{subfigure}
  \qquad
  \begin{subfigure}[t]{0.2\textwidth}
    \centering
    \captionsetup{justification=centering}
    \includegraphics[width=1.73in,]{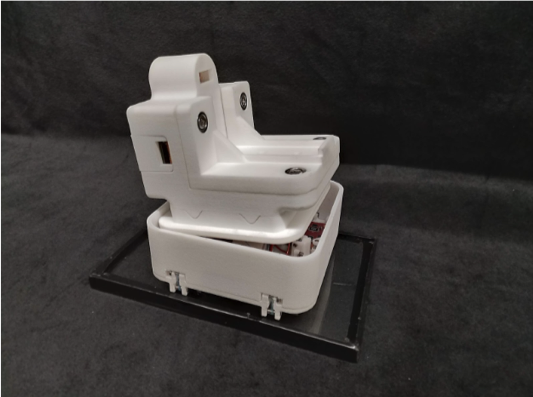}
    \caption{}
    \label{fig:design_iteration3}
  \end{subfigure}
  \qquad
    \begin{subfigure}[t]{0.2\textwidth}
    \centering
    \captionsetup{justification=centering}
    \includegraphics[width=1.72in,]{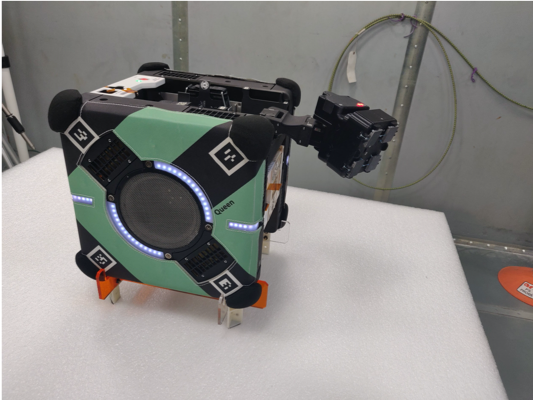}
    \caption{}
    \label{fig:design_iteration4}
  \end{subfigure}
  \caption{The flat-surface, gecko-adhesive gripper design was designed through an iterative design process. The design started with a laser-cut, hand-operated tendon, proof of concept prototype (a). Then electronics were designed and tested on a breadboard, and the tendons were controlled with servos (b). A gripper identical to the final design was printed using cheaper material to check the tolerances, fit, and interface with Astrobee (c). Then the final flight gripper was printed in ULTEM and went through rigorous pre-flight testing (d).}
\label{fig:design_iterations}
\end{figure*}

At a high-level, the flight gripper design was developed with an eye towards manipulating and grasping payloads in micro-gravity environments with assistive free-flying robots. 
Figure~\ref{fig:design_iterations} illustrates the evolution of the gripper design from a prototype to flight model. 
All components of the gecko-inspired adhesive gripper fit within the payload stowage volume of the Astrobee Gripper. 
When mated to the perching arm, the combined arm/gripper assembly can occupy any payload bay of an Astrobee free-flyer, remaining recessed such that any collisions will be to the Astrobee bumpers.

\subsection{Mechanical Design}

As shown in~\cref{fig:gripper_section}, the flight gripper is divided into four sub-assemblies: the opposed adhesive tiles, palm assembly, wrist assembly, and base assembly. 
The opposed adhesive tiles provide the adhesion necessary for perching operations. 
The palm assembly houses these opposed adhesive tiles as well as the servos, tendons, and electronics necessary for setup and control of these tiles. 
The wrist assembly connects to the palm assembly through a ball joint to allow for compliance during grasping and controls this degree of freedom with a servo and six tendons. 
Finally, the base assembly contains the gripper avionics used to control gripper functionality and to communicate with the Astrobee perching arm.

\subsubsection{Opposed Adhesive Tiles}
Two tiles with gecko-inspired adhesives are oriented in opposition to form one gripper pair. 
They are held in place by the combination of a bias spring in-line with the tiles and compliance springs behind each tile.
A Spectra (UHMWPE) wire connects the two tiles and the load tendon is tied to the middle of this inter-tile tendon.
When the two tiles are in contact with the surface, the load tendon is pulled, which shears and then applies a normal load to the engaged adhesives.
The tiles can be easily released from the surface by removing tension from the load tendon and pulling gently on the release tendons~\cite{Hawkes2015}.

\begin{figure}[h]
\centering
\includegraphics[width=.5\textwidth]{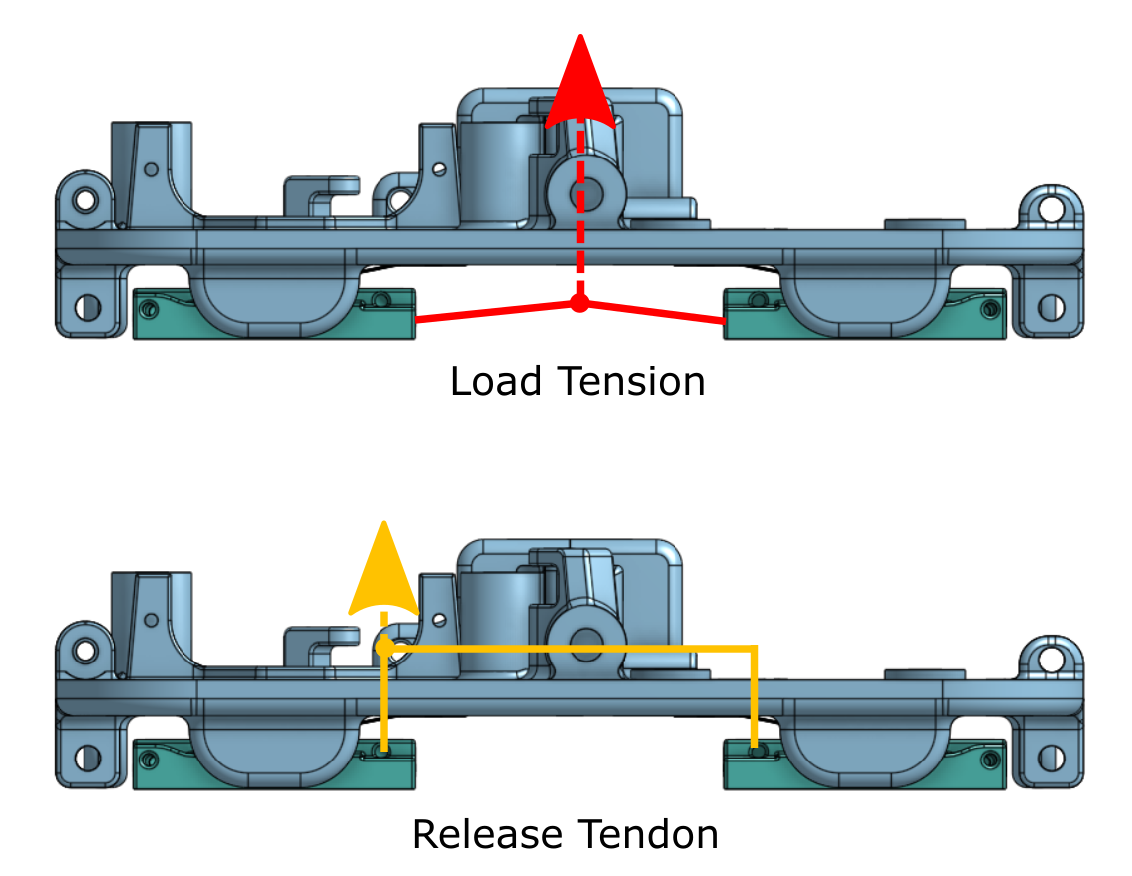}
\caption{The load tendon (highlighted in red) is tied to the middle of the tendon connecting the two opposed adhesive tiles. Controlled by a servo, the load tendon pulls the tiles first in the shear direction and then the normal, activating the adhesion. The release tendons (highlighted in gold) are tied to each corner of the adhesive tiles, introducing a local peeling moment during the release of the adhesives.}
\label{fig:wrist_tendon}
\end{figure}

This loading and unloading sequence is shown in~\cref{fig:loading_sequence}. 
The grey rectangles are the adhesive tiles, each connected to another by the tile tendon. 
To maintain initial tension in the tile tendon, tensioning tendons are attached to the outer edges of each adhesive tile. 
In this design, one pre-tension tendon is tied to a fixed point, while the other has a spring in-line to provide tension. 
A load tendon is tied to the middle of the tile tendon such that when it is pulled, it loads the tiles in both shear (inwards) and normal (upwards), generating useful grasp forces. 
Release tendons are tied to the edges of the tiles. 
When these release tendons are pulled, they apply a peeling moment to the tiles and allow for nearly zero-force release from the surface.
The locations of these tendons are shown in~\cref{fig:wrist_tendon}.

There are four tiles total on the gripper, which are divided into pairs.
The two pairs are independent from each other without any load sharing.
The total adhesion required in a micro-gravity environment to perch an Astrobee robot on the wall is well below the normal adhesion limit of one pair of adhesives~\cite{Day2013}, and thus the redundancy of two independent tiles are favored over higher combined total adhesion.
Due to flight safety requirements for operating on-board the ISS, a novel manufacturing procedure for wedge-shaped gecko-inspired adhesives was used with the gecko-inspired adhesives directly cast onto the front face of 3D-printed tiles made of AlSi10Mg.

Under optimal conditions when the grasp surface is perfectly smooth, flat, and clean, the maximum normal adhesive force is no greater than \SI{20}{\newton} for one pair of the adhesives~\cite{Day2013}.
In practice, the actual achieved grasp force is expected to be lower and depends on the surface roughness and cleanliness.
In case of emergency, forcible removal of the tiles will result in no residue or damage to the grasped surface and is unlikely to result in damage to the gripper.

\begin{figure}[h]
\centering
\includegraphics[width=.5\textwidth]{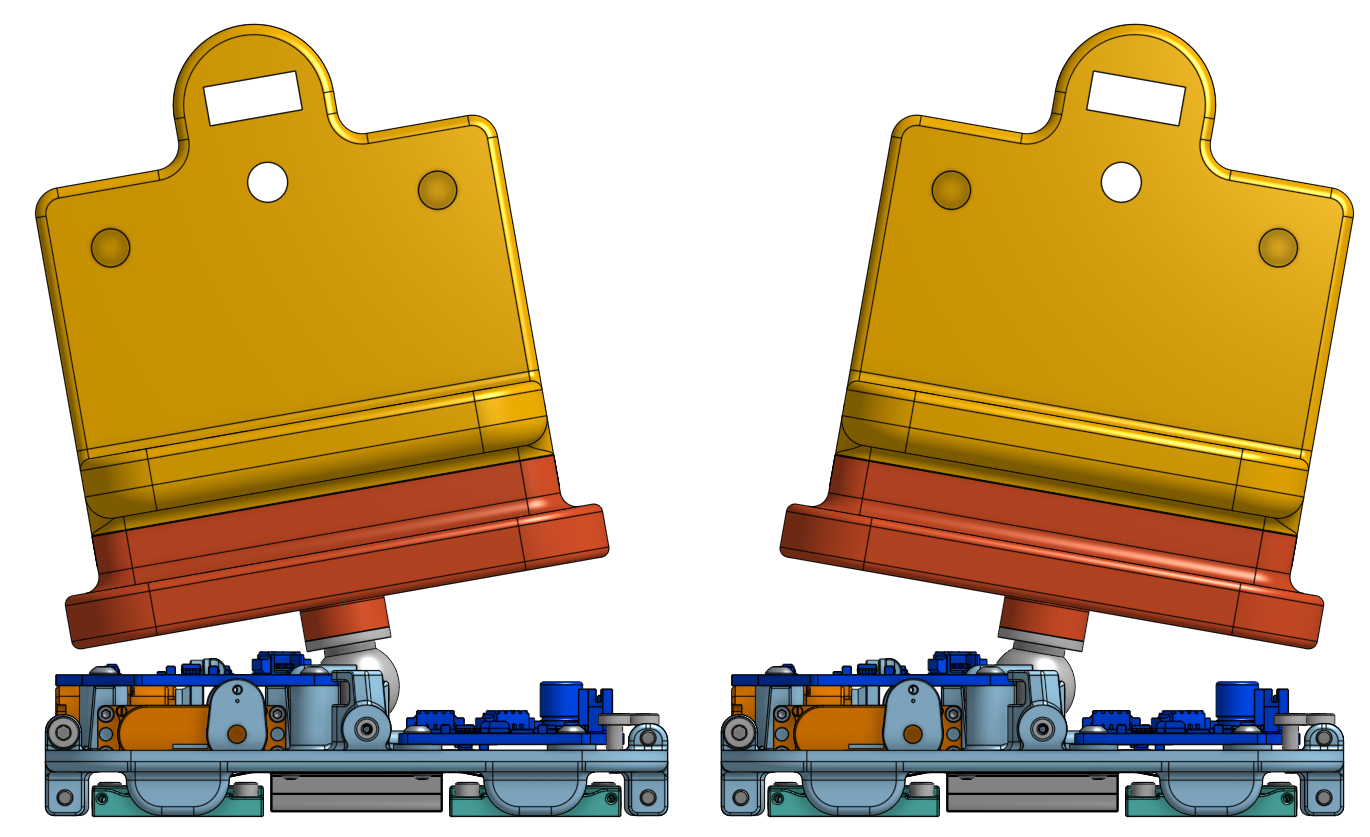}
\caption{The built-in 10 degrees of rotational compliance from the wrist joint.}
\label{fig:pitch}
\end{figure}

\begin{figure}[h]
\centering
\includegraphics[width=.5\textwidth]{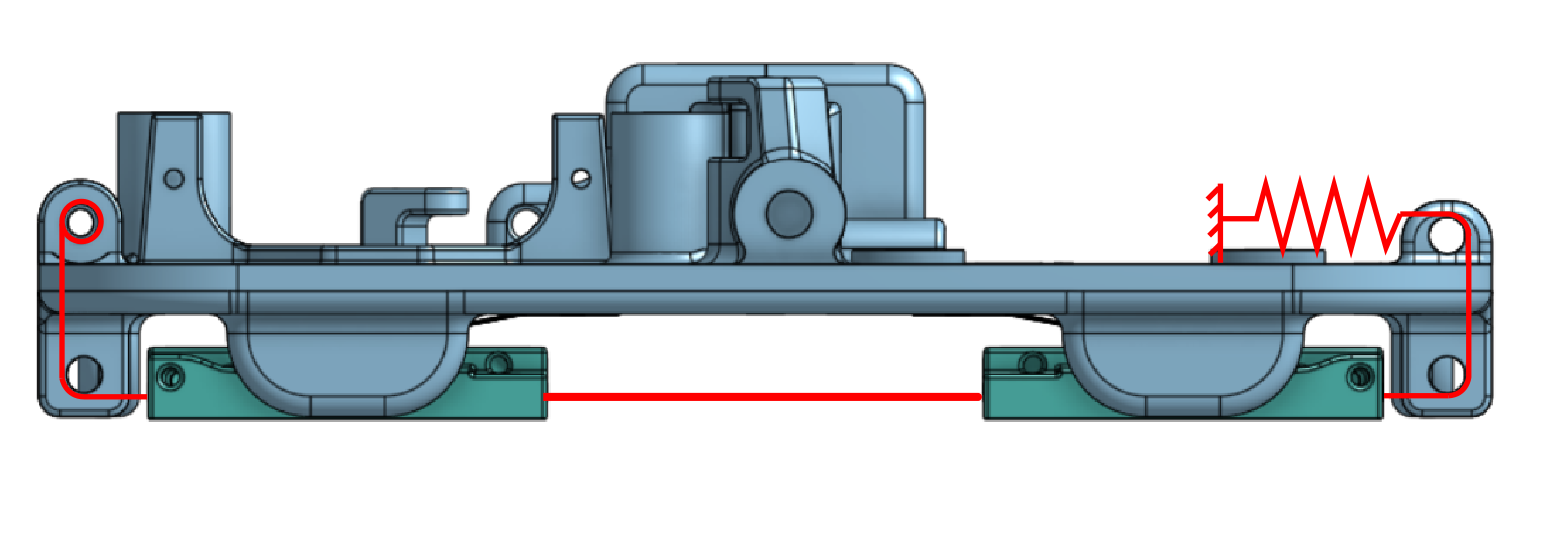}
\caption{The tendons (highlighted in red) are routed through a series of dowel pins, fixed on one side and tied to a spring on the other side to apply the necessary preload force.}
\label{fig:tendon_route}
\end{figure}

\subsubsection{Palm Assembly}

The palm  orients and controls the opposed adhesive tiles to enable grasping operations. 
For proper functionality, the linking tendon must be held in tension when not actively grasping and tendon loops are attached to the outer edges of each tile to accomplish this.
The loops from the tiles on the left in~\cref{fig:tendon_route} are routed to the inner side of the palm around a stainless steel dowel pin and fixed in place.

The loops from the tiles to the right are also routed around the dowel pin to the inside of the gripper, but are attached to an extension spring provide the necessary preload.
This tendon routing system is illustrated in~\cref{fig:wrist_tendon}.

\begin{figure}[h]
\centering
\includegraphics[width=.5\textwidth]{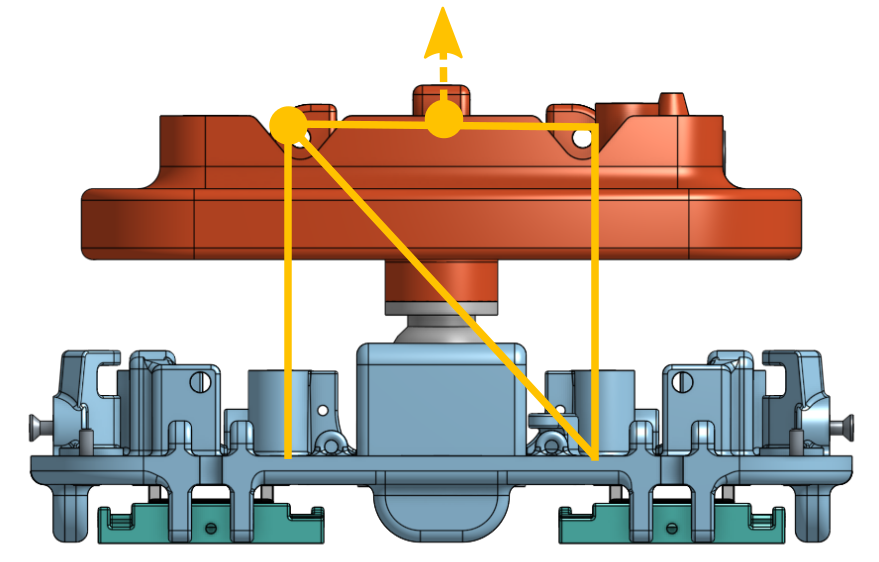}
\caption{Six tendons (highlighted in gold) are tied in this configuration to lock the pitch, yaw, and roll of the joint.}
\label{fig:rotate_tendon}
\end{figure}

Each tendon is kept taut for functionality and is constrained at intervals frequent enough to prevent potential internal entanglement.
The two load tensions are controlled through 2 hobby servos (DS65K, MKS Servos) and the release tendons are controlled with a single common servo (DS75K, MKS Servos).
The two kinds of servos are identical in specifications and only differ in the external form factor.

At the center of the palm is the housing for a ball joint on the wrist.
This is to provide compliance in the rotational direction with respect to the Astrobee during grasping.
This allows for small angular misalignment when approaching grasp target.
The palm also contains a small cutout used to house the time-of-flight distance sensor.

\subsubsection{Wrist Assembly}

The wrist provides compliance to account for slight misalignment between the Astrobee and the target perching surface.
The ball joint mates to the housing in the palm subassembly, and this provides approximately \SI{10}{\degree} of rotation in all three rotational directions.
There are three pairs of tendons routed to the servo (DS75K, MKS Servos), each acting to lock the yaw, pitch, and roll of the joint.
This is illustrated in~\cref{fig:rotate_tendon}.
When the servo  is activated, tension in the tendons locks the gripper in the neutral position, with no angular deviation between the palm and wrist components.
When the servo is deactivated, the palm is free to rotate with respect to the wrist within the available \SI{10}{\degree} of travel.
This is illustrated in~\cref{fig:pitch}.

The wrist joint is not backdrivable.
However, the joint has a small range of motion and the shroud reduces the possibility of pinch points and entanglement due to this motion.
In case of needing to move Astrobee or the perching arm specifically, this can be accomplished by back-driving the perching arm joints.

\subsubsection{Base Assembly}

The base assembly houses the microcontroller and the PCB with the electrical interface to the Astrobee perching arm.
A mechanical interface for connecting to the Astrobee perching arm is provided by the gripper adaptor using four captive screws on the adaptor.
This allows the overall assembly to act as a modular drop-in replacement for the existing claw gripper.





\subsection{Avionics Design}
\label{subsec:avionics}
\begin{figure}[t]
\centering
\includegraphics[width=.5\textwidth]{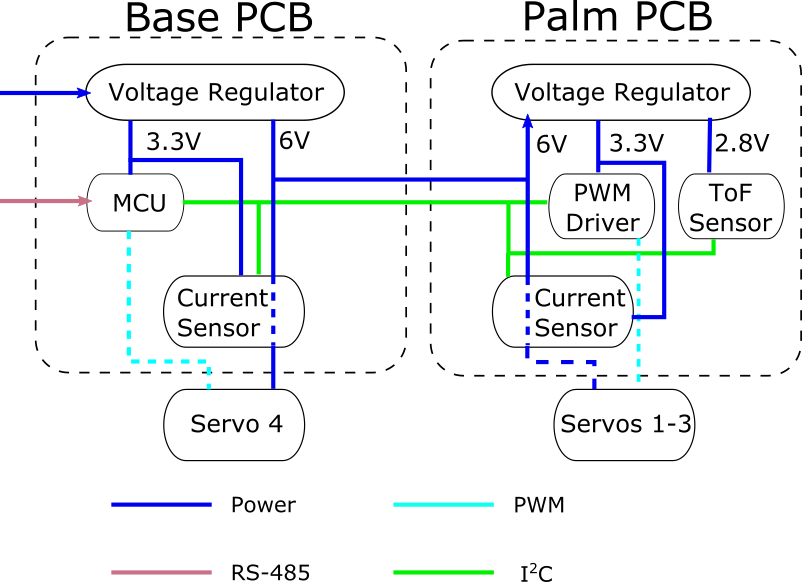}
\caption{Block diagram of the avionics system for the flight gripper. The gripper avionics are connected to the same power and data lines as the joint servo motors on the perching arm. The primary gripper avionics components include the three servos, sensor suite, and microcontroller.}
\label{fig:block_diagram}
\end{figure}

The gripper avionics stack consists of a microcontroller, a base PCB, palm PCB, four servos, and on-board sensors.
Figure~\ref{fig:block_diagram} shows a full block diagram of the avionics stack.

The microcontroller used for control of gripper functions is a COTS PRJC Teensy 3.2, which is a \SI{3.3}{\volt}, 32-bit ARM processor on a breakout board. 
The gripper firmware was written using the Teensyduino variant of the Arduino IDE and can be found at~\url{https://github.com/StanfordASL/astrobee_gripper}.

The base PCB of the gripper serves as the communication hub between the Astrobee and the \pac{} board.
As the gecko gripper shares the same power bus and RS-485 data lines with the shoulder and wrist motors (Dynamixel brand servos) on the Astrobee perching arm, the gripper simply implements the same Dynamixel-defined communication protocol so that it can be controlled from the same bus.
A 10-pin connector links the base PCB with a matching connector on the distal link on the Astrobee perching arm and provides power and ground capable of supplying \SI{3}{\ampere} at \SI{11}{\volt}.

On-board sensing consists of four DC current sensors (INA219, Texas Instruments) and one time-of-flight sensor (VL6180X, STMicroelectronics). 
The four current sensors monitor the current draw of each of the four servos during operation and are used to estimate of the tension in each of the tendons. 
For data collection and storage purposes, a microSD card mounted on the base PCB stores the time-of-flight and current sensor measurements, and an on-orbit ``slow-drip'' procedure is used to send recorded data to the operator without removing the microSD card.

A time-of-flight (\tof{}) sensor measuring laser-based distance measurements is used to estimate the distance between the gripper and the targeted perching surface. The measurement range for the VL6180X is from \SIrange{5}{100}{\milli\meter}. 
We estimate the time-to-contact between the gripper and perching surface using \tof{} measurements and automatically engage the gripper adhesive in an open-loop fashion. This is done as the gecko gripper requires pose estimates at a centimeter resolution, which is greater precision than the on-board Astrobee filter can provide.
First, a distance threshold of \SI{40}{\milli\meter} is used to detect when the perching surface is within the perching range of the gripper and the gripper velocity estimated by finite differences of the \tof{} distance measurements. 
We assume that the Astrobee controller approaches the perching surface such that the gripper surface and perching surface normals are in alignment, and consequently we assume a constant-direction velocity. 
Further, as the Astrobee thrusters provide a maximum of only \SI{10}{\centi\meter\per\second\squared} of thrust along any axis, we also assume a constant velocity profile over this final \SI{40}{\milli\meter} distance and set the time-to-contact before the adhesives are automatically triggered. 
In practice, we found that an additional delay of \SI{250}{\milli\second} was necessary for reliably computing the time-to-contact, and future work includes introducing feedback on the \tof{} measurements for closed-loop control.

\begin{figure*}[t!]
    \centering
    \hfill
    \begin{subfigure}[t]{0.3\textwidth}
        \centering
        \includegraphics[width=\textwidth]{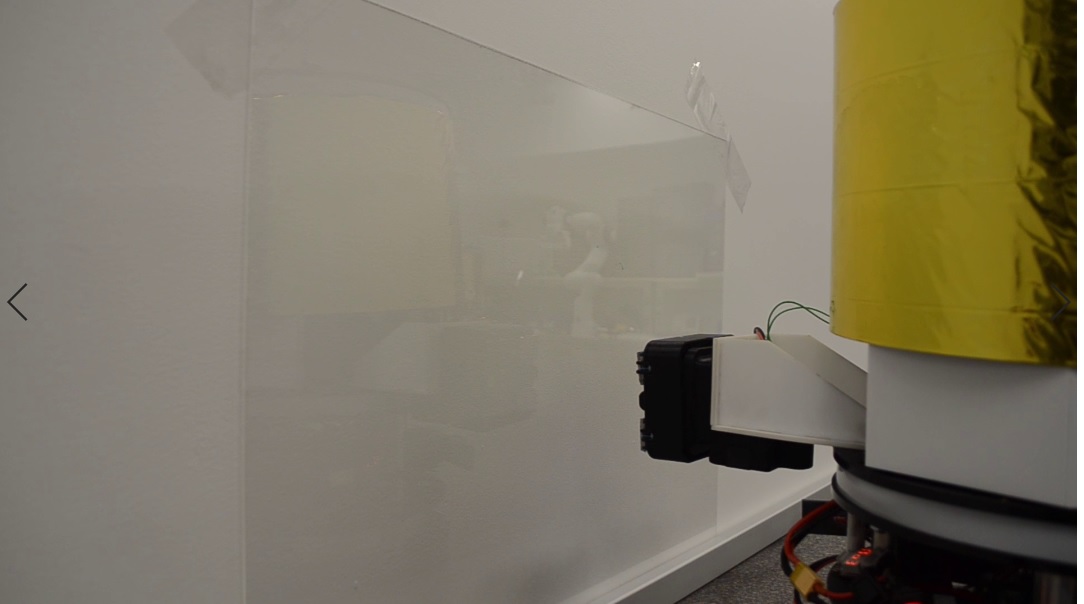}
        \caption{Flight gecko-gripper on the Stanford free-flyer with an acrylic surface as the perching target.}
        \label{fig:stanford_testing_1}
    \end{subfigure}
    \hfill
    \begin{subfigure}[t]{0.3\textwidth}
        \centering
        \includegraphics[width=\textwidth]{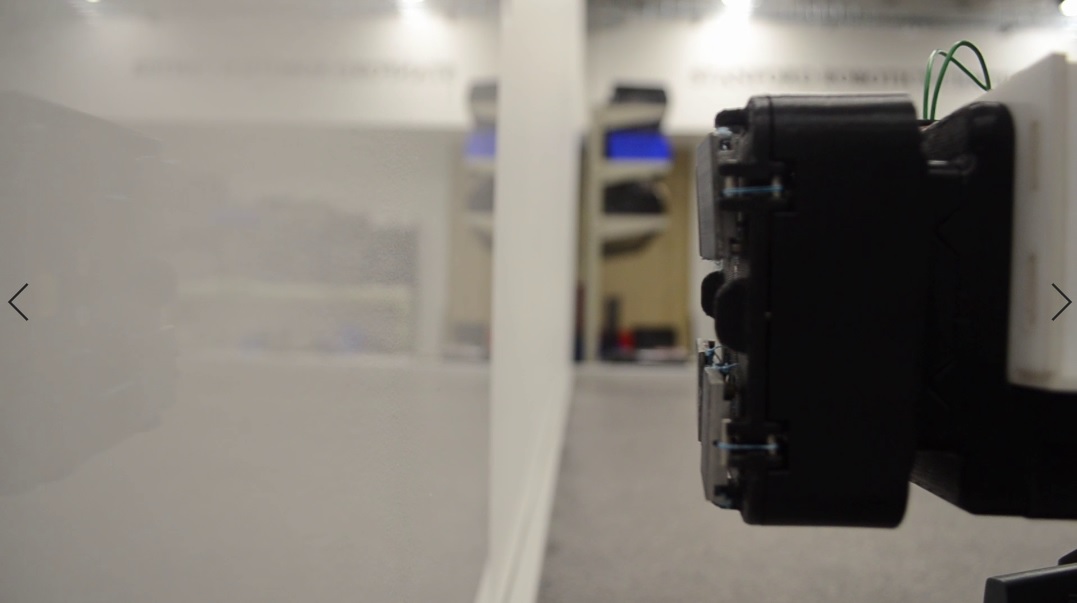}
        \caption{The opposed adhesive tile pairs can be seen on the gripper.}
        \label{fig:stanford_testing_2}
    \end{subfigure}
    \hfill
    \begin{subfigure}[t]{0.3\textwidth}
        \centering
        \includegraphics[width=\textwidth]{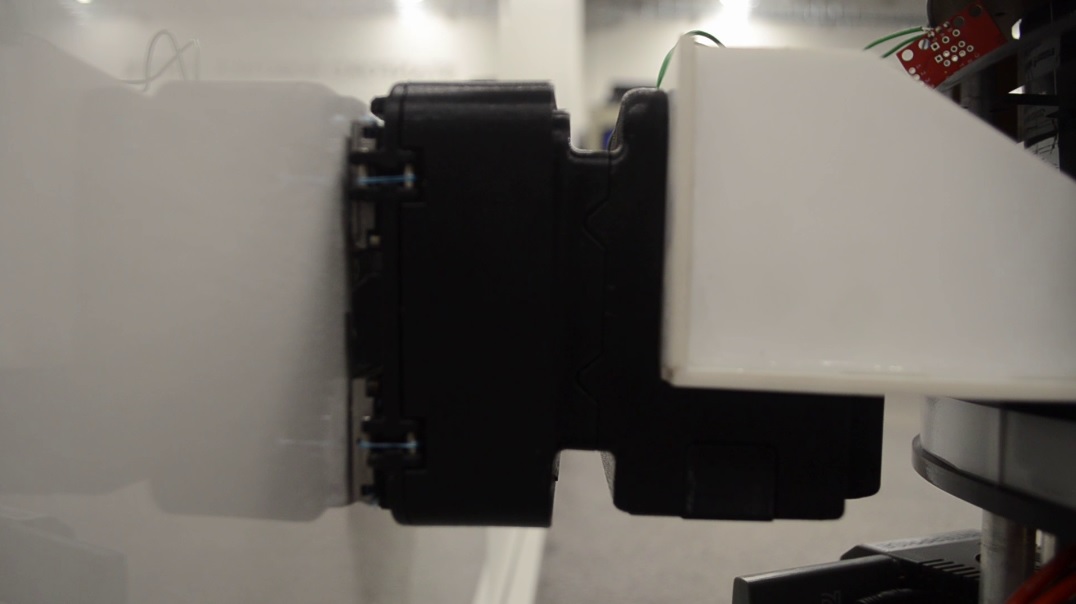}
        \caption{Time-to-contact with the perching surface was estimated using \tof{} measurements to trigger the grasp sequence.}
        \label{fig:stanford_testing_3}
    \end{subfigure}
    \hfill
\caption{Perching experiments with a flight gripper unit were conducted on the Stanford spacecraft robotic test bed~\cite{MacPhersonHockmanEtAl2017}. }
\label{fig:stanford_granite_testing}
\end{figure*}

\subsection{Software}
Astrobee has three ARM processors, of which the low-level processor (LLP) and mid-level processor (MLP) both run Linux, and the high-level processor (HLP) runs Android. 
The HLP serves as the hub to run guest science applications and programs and the MLP runs the primary navigation and planning algorithms on-board. 
Finally, the LLP manages low-level thruster allocation commands and the EKF.
For integrating the gecko gripper firmware, we modify firmware on both the \mlp{} and \pac{} to be able to issue commands and receive data packets from the gecko griper.

The \pac{} serves as the intermediary between the \mlp{} and Astrobee shoulder and arm motors. 
The firmware on the \pac{}'s dsPIC33EP microprocessor is designed to provide extensible communications with both the \mlp{} over an internally-defined serial protocol as well as any device on the arm's RS-485 bus implementing the Dynamixel servo protocol (a reasonably generic read-write interface to register addresses).  Both provide framing, checksumming, and packet acknowledgement for reliable communication, with numeric command identifiers (register addresses) specifying different actions or status inquiries.  Therefore, we augmented the firmware to handle new \mlp{} packet types that issue corresponding commands to the gecko gripper processor acting as a virtual servo.  In this way, we provide a first demonstration of the relative simplicity and modularity of implementing alternative grippers for the Astrobee perching arm.

For the \mlp{}, we take advantage of the modular ROS framework and use a high-level interface for integration. 
A ROS node on the Astrobee flight software communicates with the \pac{} to control the perching arm joint positions and to open/close the Astrobee gripper. 
We simply make use of the custom command message definition for this node to accommodate the additional commands defined for the gecko gripper. The full set of commands and write registers for the gecko gripper are detailed in Table~\ref{table:commands}.

\subsection{Communications}
On-orbit operation of the gecko gripper on Astrobee is to be managed through the use of the NASA Ground Data System.
During these experiments, ground station operators will enable gripper functionality and send commands to a Guest Science application running on Astrobee. This Guest Science application runs on the Astrobee HLP and serves as an interface with ground station operators to parse and interpret the commands received by the operator~\cite{Vargas2018}.
Existing mobility modules developed for Guest Science users provide setpoint commands for the Astrobee robots. 
Additional functionality developed for the ground station operators allow for setting the gecko gripper into autonomous mode for perching experiments and read/write commands to record science data on the microSD card on the gripper. 
Finally, these commands are parsed by the MLP, \pac{}, and gecko gripper microcontroller as described in~\ref{subsec:avionics}.

\section{Testing}
\subsection{Ground Testing}
The development of a flight-qualified gripper design also entailed extensive validation and safety testing. 
This testing included:
\begin{enumerate}
    \item Vibration testing: structural integrity was tested by conducting vibration testing at a load factor of 12\,G in each axis for launch with the payload stowed in bubble wrap to mimic in-flight packaging.
    \item Electromagnetic interference (EMI) testing: An integrated EMI test with an Astrobee unit was performed at NASA Johnson Space Center (JSC). An exceedance was identified around a similar frequency as the base Astrobee unit. A Tailoring/Interpretation Agreement (TIA) was approved for the grippers with the same operational limits as the base Astrobee.
    \item Acoustic testing: To comply with the noise limits on the ISS, an acoustics test was performed on the two flight grippers at JSC in an acoustic room. It was measured that neither gripper exceed the overall noise level of an Astrobee during nominal operation, and appropriate limits on operation time were applied based on the overall noise level.
    \item Thermal testing: Temperatures were measured for a single servo operating in an off-nominal, continuously stalled condition---the worst-case single-failure thermal load.  Temperature measurements in a convection-restricted enclosure verified that no point on the exterior of the gripper exceeds the maximum permissible touch temperature.
\end{enumerate}

To verify the adhesion performance of the grippers before packaging, we performed simple pull tests of the gripper on a clean, flat sheet of acrylic.
We first attached the gripper to the acrylic sheet.
On the other side of the acrylic sheet we attached a force gauge. While holding the gripper still, we pulled the force gauge until the acrylic sheet detached from the gripper and the maximum force recorded.
Only five trials were performed as a functionality check on the two flight grippers since pulling off the adhesives without disengaging them would cause minor deterioration on adhesives over time.
The mean pull-off force was measured at 10.4N and 11.2N for the two grippers, with all trials fall within 10 percent of the mean.

Finally, preliminary testing to verify the functionality of the gecko gripper was conducted on a 3-\dof{} granite table test bed with both the Astrobee robot and Stanford free-flyer robots. 
As shown in~\cref{fig:granite_table}, the test setup with the Astrobee robot on the granite table at NASA Ames Research Center consisted of placing a smooth acrylic sheet as the target surface for the deployed perching arm. 
The controller used was a simple PD controller with a target waypoint behind the target surface such that the gripper makes contact with the surface with non-zero momentum.
Similar testing was conducted on the Stanford spacecraft robotic test bed as shown in~\cref{fig:stanford_granite_testing}.

\subsection{Flight Experiments}
\label{subsec:flight_experiments}
Two flight units for the gecko adhesive gripper were launched to the ISS in July 2019, and they are currently scheduled for on orbit experiments with the Astrobee robot in 2021. 
Prior to commencing perching experiments, the Astrobee gripper will be detached and replaced with the gecko adhesive gripper. 
The perching experiments will consist of two phases. 
First, manual adhesion testing will be conducted by an astronaut guiding the Astrobee unit towards a polished aluminum test surface that was also launched to verify basic gripper functionality and operations. 
Thereafter, the second phase entails conducting autonomous perching experiments with a designated ISS bulkhead or panel surface as the perching target. 
The Astrobee free-flyer will use a PID controller to approach the target in an uncluttered environment with automatic grasping mode enabled on the gripper. 
In both phases, ground station operators will be in-the-loop to command the gripper between perching trials. 
After experiments have concluded, science data recorded on the gripper microSD card will be streamed back to the operators for post-processing.

\section{Future Work}
In this work, we presented the design and development of a flight gripper for the Astrobee free-flying robot. 
The design drivers for this gripper included flight safety considerations for operating inside the ISS and constraints driven by having to integrate with the Astrobee perching arm. 
As detailed in~\cref{subsec:flight_experiments}, two flight units of the gecko gripper were launched in 2019 and are currently scheduled for in-flight experiments with the Astrobee robot in 2021. 
Through these experiments, we hope to validate the efficacy and robustness of the developed gripper design for carrying out tasks involving dexterous grasping and manipulation. 
Through lessons learned during these experiments, we seek to enable a future in which astronauts and assistive free-flying robots can safely and effectively work alongside each other.

\subsection*{Acknowledgements}
This work was supported in part by NASA under the Space Technology Research Fellowship, Grants NNX16AM46H and NNX16AM78H, and the NASA Early Stage Innovations Grant NNX16AD19G. The authors wish to thank Andrew Bylard, Jonathan Barlow, Jose Benavides, Maria Bualat, Tyler Dorval, Lorenzo Fl{\"u}ckiger, Terry Fong, and Trey Smith for their discussions during the course of this work.

\bibliographystyle{isairas}
{\small
 \bibliography{ASL_papers,main,paper}

\newcommand{\noopsort}[1]{} \newcommand{\printfirst}[2]{#1}
  \newcommand{\singleletter}[1]{#1} \newcommand{\switchargs}[2]{#2#1}
\begin{thebibliography}{10}
\expandafter\ifx\csname url\endcsname\relax
  \def\url#1{{\tt #1}}\fi
\expandafter\ifx\csname urlprefix\endcsname\relax\def\urlprefix{URL }\fi
\expandafter\ifx\csname urlstyle\endcsname\relax
  \expandafter\ifx\csname doi\endcsname\relax
  \def\doi#1{doi:\discretionary{}{}{}#1}\fi \else
  \expandafter\ifx\csname doi\endcsname\relax
  \def\doi{doi:\discretionary{}{}{}\begingroup \urlstyle{rm}\Url}\fi \fi

\bibitem{EstradaHockmanEtAl2016}
Estrada MA, Hockman B, Bylard A, Hawkes EW, Cutkosky MR and Pavone M (2016)
  \hspace{0pt}Free-Flyer Acquisition of Spinning Objects with Gecko-Inspired
  Adhesives.
\newblock In: {\it {Proc.\ IEEE Conf.\ on Robotics and Automation}\/}.

\bibitem{Vargas2018}
Vargas AM, Ruiz RG, Wofford P, Kumar V, {Van}~{Ross} B, Katterhagen A, Barlow
  J, Fl{\"u}ckiger L, Benavides J, Smith T and Bualat M (2018)
  \hspace{0pt}{Astrobee}: Current Status and Future Use as an International
  Research Platform.
\newblock In: {\it {Int.\ Astronautical Congress}\/}.

\bibitem{MacPhersonHockmanEtAl2017}
MacPherson R, Hockman B, Bylard A, Estrada MA, Cutkosky MR and Pavone M (2017)
  \hspace{0pt}Trajectory Optimization for Dynamic Grasping in Space using
  Adhesive Grippers.
\newblock In: {\it {Field and Service Robotics}\/}.

\bibitem{Flores-AbadMaEtAl2014}
{Flores-Abad} A, Ma O, Pham K and Ulrich S (2014) \hspace{0pt}A review of space
  robotics technologies for on-orbit servicing.
\newblock In: {\it {Progress in Aerospace Sciences}\/}, 68(1):pp.1--26.

\bibitem{BylardMacPhersonEtAl2017}
Bylard A, MacPherson R, Hockman B, Cutkosky MR and Pavone M (2017)
  \hspace{0pt}Robust Capture and Deorbit of Rocket Body Debris Using
  Controllable Dry Adhesion.
\newblock In: {\it {IEEE Aerospace Conference}\/}.

\bibitem{BualatSmithEtAl2018}
Bualat MG, Smith T, Fong TW, Smith EE, Wheeler DW and {The Astrobee Team}
  (2018) \hspace{0pt}{Astrobee:} {A} New Tool for {ISS} Operations.
\newblock In: {\it {Int.\ Conf.\ on Space Operations (SpaceOps)}\/}.

\bibitem{Enright2004}
Enright J, Hilstad M, Saenz-Otero A and Miller D (2004) \hspace{0pt}The
  {SPHERES} {Guest} {Science} {Program}: Collaborative Science on the {ISS}.
\newblock In: {\it {IEEE Aerospace Conference}\/}.

\bibitem{SmithBarlowEtAl2016}
Smith T, Barlow J, Bualat M, Fong T, Provencher C, Sanchez H and Smith E (2016)
  \hspace{0pt}{Astrobee:} {A} New Platform for Free-Flying Robotics on the
  {International Space Station}.
\newblock In: {\it {Int.\ Symp.\ on Artificial Intelligence, Robotics and
  Automation in Space}\/}.

\bibitem{Park2017}
Park IW, Smith T, Sanchez H, Wong SW, Piacenza P and Ciocarlie M (2017)
  \hspace{0pt}Developing a {3-DOF} compliant perching arm for a free-flying
  robot on the {International} {Space} {Station}.
\newblock In: {\it {IEEE/ASME Int.\ Conf.\ on Advanced Intelligent
  Mechatronics}\/}.

\bibitem{Autumn2006}
Autumn K, Dittmore A, Santos D, Spenko M and Cutkosky M (2006)
  \hspace{0pt}Frictional adhesion: a new angle on gecko attachment.
\newblock In: {\it {Journal of Experimental Biology}\/}.

\bibitem{Song2017}
Song S, Drotlef DM, Majidi C and Sitti M (2017) \hspace{0pt}Controllable load
  sharing for soft adhesive interfaces on three-dimensional surfaces.
\newblock In: {\it {Proceedings of the National Academy of Sciences}\/}.

\bibitem{Trentlage2018}
Trentlage C and Stoll E (2018) \hspace{0pt}A Biomimetic Docking Mechanism for
  Controlling Uncooperative Satellites on the {ELISSA} Free-Floating
  Laboratory.
\newblock In: {\it {IEEE/ASME Int.\ Conf.\ on Advanced Robotics \&
  Mechatronics}\/}.

\bibitem{Trentlage2018a}
Trentlage C, Hensel R, Holzbauer R, Stoll E, Arzt E and Makaya A (2018)
  \hspace{0pt}Development of Gecko-Inspired Adhesive Materials for Space
  Applications.
\newblock In: {\it {Int.\ Astronautical Congress}\/}.

\bibitem{JiangHawkesEtAl2017}
Jiang H, Hawkes EW, Fuller C, Estrada MA, Suresh SA, Abcouwer N, Han AK, Wang
  S, Ploch CJ, Parness A and Cutkosky MR (2017) \hspace{0pt}A robotic device
  using gecko-inspired adhesives can grasp and manipulate large objects in
  microgravity.
\newblock In: {\it {Science Robotics}\/}, 2(7):pp.1--11.

\bibitem{Parness2017a}
Parness A (2017) \hspace{0pt}Testing Gecko-Like Adhesives Aboard the
  {International} {Space} {Station}.
\newblock In: {\it {AIAA SPACE and Astronautics Forum \& Exposition}\/}.

\bibitem{Hawkes2015}
Hawkes EW, Jiang H and Cutkosky MR (2015) \hspace{0pt}Three-dimensional dynamic
  surface grasping with dry adhesion.
\newblock In: {\it {Int.\ Journal of Robotics Research}\/}.

\bibitem{Day2013}
Day P, Eason EV, Esparza N, Christensen D and Cutkosky M (2013)
  \hspace{0pt}Microwedge Machining for the Manufacture of Directional Dry
  Adhesives.
\newblock In: {\it {ASME Journal of Micro and Nano-Manufacturing}\/}.

\end{thebibliography}
}

\newpage

\section*{Appendix}
\begin{table}[h!]
\caption{List of commands defined for the flight gripper.}
\rowcolors{1}{}{lightgray}
\begin{tabular}{@{} m{2cm}  m{5.5cm} @{} }
    \hline
    \textbf{Command Name} &  \textbf{Description}\\
    \hline
    OPEN & Disengage the adhesives by releasing the tension in the load tendons and then pulsing the release tendon. Disable automatic grasp mode. After a timed delay, unlock the wrist. Ideally, this results in the wrist remaining near center in a zero-gravity environment.\\
    CLOSE & Engage the adhesives, and then after a timed delay, smoothly lock the wrist.\\
     TOGGLE AUTO & Toggle whether automatic grasp mode is enabled or disabled.\\
     MARK & Starts data logging for a new experiment number. Experiment 0 is reserved word to indicate termination of logging.\\
    ENGAGE & Engage the adhesives, while taking no further action.\\
    DISENGAGE & Disengage the adhesive by releasing the tension in the load tendons, and then pulsing the release tendon.\\
    LOCK & Slowly lock the wrist \dof{} by smoothly tightening the wrist tendon.\\
    UNLOCK & Smoothly unlock the wrist \dof{} by smoothly loosening the wrist tendon.\\
    ENABLE AUTO & Enable automatic grasp mode.\\
    DISABLE AUTO & Disable automatic grasp mode, without affecting the state of any active grasps.\\
    SET DELAY & Set the delay, in milliseconds, used during grasping to close the gripper.\\
    STATUS & Two-byte register containing bits for adhesive, wrist, automatic mode, and experiment-in-progress statuses. Further bits may be added as needed.\\
    RECORD & Register address from which reading 35 bytes will return the current record in the open experiment file. If no file is open, returns zeros.\\
    OPEN EXP & Open a specific experiment file in preparation for feeding data back to the \pac{}.\\
    CLOSE EXP & Close the current experiment file.\\
    NEXT RECORD & Seek to the next record from the open experiment file. Parameter indicates the number of records to seek by.\\
   \hline
\end{tabular}
\label{table:commands}
\end{table}

\end{document}